\documentclass[lettersize,journal,onecolumn,draftclsnofoot]{IEEEtran}

\usepackage{lineno}
\usepackage{amsmath,amsfonts}
\usepackage[english]{babel}
\usepackage{xcolor}
\usepackage{graphicx}
\usepackage{xcolor}
\usepackage{colortbl}
\usepackage{float}
\usepackage{cite}

\usepackage{hyperref}
\hypersetup{
    colorlinks=true,
    linkcolor=blue,
    citecolor=blue,
    filecolor=magenta,
    urlcolor=cyan,
    pdfpagemode=FullScreen,
}

\usepackage{algorithm}
\usepackage{algorithmic}
\usepackage{afterpage}
\usepackage{array}
\usepackage[caption=false,font=normalsize,labelfont=sf,textfont=sf]{subfig} 
\usepackage{xcolor}
\usepackage{longtable}
\usepackage{booktabs}
\usepackage{textcomp}
\usepackage{stfloats}
\usepackage{url}
\usepackage{verbatim}
\usepackage{graphicx}
\usepackage{tikz}
\usepackage{multirow}
\usepackage{pdflscape}
\usepackage{siunitx}
\usepackage{lscape}
\usepackage{booktabs}
\usepackage{multirow}
\usepackage{pdflscape} 
\newcolumntype{C}[1]{>{\centering\arraybackslash}m{#1}}

\addto\extrasenglish{%
}

\begin{document}

\title{Is Self-Pretraining really useful to improve diagnosis in medical Time Series?}

\author{%
  Omar Coser$^{1,2}$, Antonio Orvieto$^{4,5}$, Paolo Soda$^{1,3}$, Loredana Zollo$^{2}$%
  \thanks{$^{*}$ Corresponding author: O.\ Coser (email: omarcoser10@gmail.com).}%
  \thanks{$^{1}$Unit of Artificial Intelligence \& Computer Systems, Università Campus Bio-Medico di Roma, Via Álvaro del Portillo, 21, Rome, 00128, Italy.}%
  \thanks{$^{2}$Unit of Advanced Robotics and Human-Centered Technologies, Università Campus Bio-Medico di Roma, Via Álvaro del Portillo, 21, Rome, 00128, Italy.}%
  \thanks{$^{3}$Department of Diagnostics and Intervention, Radiation Physics, Biomedical Engineering, Umeå University, Universitetstorget, 4, Umeå, 40196, Sweden.}%
  \thanks{$^{4}$Max Planck Institute for Intelligent Systems}%
\thanks{$^{5}$ELLIS Institute Tübingen}%
}

\maketitle

\begin{abstract}
Inspired by recent evidence that transformer architectures benefit from Self-PreTraining (SPT) on long-context benchmarks, we investigate whether similar gains extend to multimodal, multivariate, and even simple univariate medical time series. Our objective is to assess the impact of SPT on the performance and scalability of transformer-based models across diverse medical applications, particularly under limited data conditions. We evaluate transformer architectures on three representative medical time-series tasks: rehabilitation robotics (Camargo dataset), stress detection (Non-EEG Stress), and Parkinson’s disease detection (Gait Parkinson’s Disease v1.0.0). Models are trained either from scratch or through SPT using four masking-based objectives designed to promote temporal and cross-modal representation learning, and we systematically vary model depth to examine how capacity interacts with pre-training benefits. Across datasets and configurations, SPT consistently improves classification accuracy by 0–6 percentage points depending on masking strategy, dataset and architecture, with gains observed not only in multivariate settings but also when models are restricted to simple univariate inputs. The improvements increase for deeper models that can better exploit the enriched temporal representations learned during pre-training. These findings indicate that SPT is a simple and general strategy that enhances transformer performance on medical time-series tasks without requiring task-specific architectural changes, supporting its potential to improve robustness and accuracy in data-limited clinical settings.
\end{abstract}


\section{Introduction}
\subsection{Time Series in the World}

Time series data are pervasive in domains such as healthcare, industrial monitoring, finance, and autonomous driving. Despite their ubiquity, developing high performance models for time-series analysis remains challenging, particularly in domains where labeled data are scarce or expensive to obtain. This limitation has made \emph{pre-training} a central paradigm for learning robust and transferable representations.

In this context, it is important to distinguish between \textbf{Self-Supervised Learning (SSL)} and \textbf{Self-PreTraining (SPT)}, two closely related yet conceptually distinct strategies for performing pre-training on time-series data. SSL typically refers to large scale pre-training performed on an external dataset that differs from the one used during downstream fine-tuning. In contrast, SPT denotes a setting in which the model is pre-trained directly on the same dataset that will later be used for fine tuning~\cite{amos2023never, coser2026towards}.

In SSL, the model is trained using automatically generated supervisory signals (e.g., forecasting, masking, or contrastive objectives) on unlabeled data, often collected from diverse sources or domains. The goal is to learn broadly transferable representations that can generalize across tasks and datasets. After this pre-training stage, the model is fine-tuned on labeled data specific to the downstream task.

In contrast, SPT exploits the unlabeled portion or alternative views of the target dataset itself. Rather than leveraging out-of-domain data, SPT focuses on in-domain pre-training to improve optimization, representation quality, and convergence specifically for the target task.

Conceptually, SSL can be viewed as \emph{out of domain representation learning}, whereas SPT corresponds to \emph{in domain pre-training}. While both approaches rely on self-supervised objectives, they differ fundamentally in data usage and generalization scope.

This distinction is particularly important in real world time series applications. In industrial or multimodal settings, large external corpora may be available, making SSL attractive. However, in highly specialized domains such as healthcare, external datasets may be limited, heterogeneous, or inaccessible due to privacy constraints. In such cases, SPT provides a practical and effective alternative by maximizing the utility of the available in domain data.

Motivated by these challenges, there has been a surge of research in self-supervised pre-training and transfer learning for time series tasks, aiming to leverage abundant unlabeled data to learn useful representations before fine-tuning on specific downstream objectives. In the following, we review recent developments in self-supervised and transfer learning frameworks for time series modeling, with particular emphasis on medical, industrial, and multimodal applications.
\paragraph{Literature Selection and Organization.}
The papers included in this review were selected through a systematic search on Google Scholar, Scopus, and Web of Science. The primary keywords used were \emph{Self-Supervised Learning in Time Series}, \emph{Self-Supervised Learning in Medical Time Series}, \emph{Pre-training in Time Series}, \emph{Pre-training in Medical Time Series}, \emph{Self-PreTraining in Time Series}, and \emph{Self-PreTraining in Medical Time Series}. The search predominantly returned works related to self-supervised learning (SSL), reflecting the maturity and rapid expansion of this research direction. 

The selected works are organized thematically to provide a structured overview of the field. We begin with general SSL approaches for time-series modeling (Section~\ref{sec:sspt_ts}), highlighting foundational pretext tasks and representation learning strategies. We then move to pre-training in clinical and multimodal contexts (Section~\ref{sec:clinical_multimodal}), where domain-specific challenges such as irregular sampling and multimodal alignment are central. Next, we discuss efforts toward scaling pre-training across heterogeneous domains (Section~\ref{sec:scaling_domains}), emphasizing large pre-trained backbones and cross-domain generalization. Finally, we provide insights from other modalities (Section~\ref{sec:other_modalities}), including vision and speech, whose methodological advances have strongly influenced time series self-supervised learning.

\subsection{Self‑Supervised Pre‑training for Time Series}
\label{sec:sspt_ts}
Early work such as Shi et al. introduced a self‑supervised pre‑training pipeline tailored for time‑series classification, using denoising and Dynamic Time Warping (DTW)–based similarity discrimination to learn transferable representations from large unlabeled corpora like UCR2015 \cite{shi2021self}.
Following this direction, Dong et al. proposed TimeSiam, a Siamese‑network framework that reconstructs masked subsequences by leveraging past temporal contexts, achieving state‑of‑the‑art results across forecasting and classification tasks \cite{dong2024timesiam}.
Similarly, Zhang et al. explored time–frequency consistency as a pretext, aligning representations in time and frequency domains through contrastive learning and showing impressive cross‑domain transfer \cite{zhang2022self}.
These methods highlight how diverse self‑supervised objectives—ranging from masked reconstruction to contrastive alignment can extract latent temporal structures without requiring labels.

The field has also benefited from comprehensive surveys that synthesize these advances.
Zhang et al. provided a taxonomy of self‑supervised learning (SSL) for time series, categorizing approaches into generative, contrastive, and adversarial paradigms and identifying key challenges such as seasonal variability and multivariate augmentation design \cite{zhang2024self}.
Liu et al. offered a domain‑specific review focused on medical time‑series, discussing data augmentations, encoder choices, and loss functions used in 43 studies for applications like ICU mortality prediction and sleep scoring \cite{liu2023self}.
Most recently, Ma et al. reviewed large pre‑trained models for time‑series mining, noting the growing adoption of transformer backbones and masked modeling strategies \cite{ma2024survey}.
These surveys underscore both the promise of pre‑training and the open research questions in designing generalizable representations.

\subsection{Pre‑training in Clinical and Multimodal Contexts}
\label{sec:clinical_multimodal}
In the medical domain, labeled data scarcity is particularly acute.
King et al. introduced a multimodal pretraining framework that aligns ICU time‑series measurements with clinical notes, using contrastive objectives and masked token prediction to build encoders that transfer effectively to mortality prediction and phenotyping \cite{king2023multimodal}.
Similarly, Raghu et al. explored contrastive pre‑training for multimodal clinical time series, integrating high‑frequency ECG signals with labs and vitals to enhance downstream detection of elevated mPAP and mortality \cite{raghu2022contrastive}.
Xu et al. proposed TransEHR, a self‑supervised transformer architecture that models asynchronous events and measurements through masked value prediction, replaced‑token detection, and event forecasting \cite{xu2023transehr}.
These works demonstrate that multimodal alignment and event‑aware architectures are critical when transferring self‑supervised representations in healthcare.

Several frameworks also address irregular or asynchronous sampling patterns, which are common in EHRs and IoT settings.
Chowdhury et al. introduced PrimeNet, combining time‑sensitive contrastive learning with duration‑based masking to better encode irregularities in multivariate time series \cite{chowdhury2023primenet}.
Malhotra et al.’s earlier TimeNet also tackled irregular data, leveraging self‑supervised reconstruction and contrastive losses to improve few‑shot transfer in clinical and sensor datasets \cite{malhotra2017timenet}.
These methods highlight the importance of domain‑specific pretext designs that account for non‑uniform temporal structures.

\subsection{Scaling Pre‑training Across Domains}
\label{sec:scaling_domains}
Beyond clinical settings, researchers have explored how large pre‑trained models can generalize across disparate domains.
Prabhakar et al. proposed LPTM, introducing adaptive segmentation during masked modeling to handle varying sampling rates and temporal dynamics in multi‑domain datasets (epidemiology, energy, traffic, and behavioral sensors) \cite{prabhakar2024large}.
Dong et al.’s SimMTM further demonstrated that series‑wise contrastive alignment combined with masked reconstruction improves performance on forecasting, classification, and imputation across domains \cite{dong2023simmtm}.
Similarly, Zhang et al. introduced UniMTS, which unifies motion time‑series with semantic text descriptions to achieve robust zero‑shot and few‑shot performance \cite{zhang2024unimts}.
These advances suggest that large unlabeled datasets, combined with carefully crafted pretext tasks, can yield universal backbones for time‑series analysis.

\subsection{Insights from Other Modalities}
\label{sec:other_modalities}
Parallel research in computer vision and speech has inspired many pre‑training strategies adopted for time series.
For instance, masked autoencoders have been widely explored for medical imaging tasks, where self‑pretraining directly on the target dataset mitigates domain shift and improves segmentation and classification \cite{zhou2023self,li2023robust,tu2022self,das2025self,rohrich2025masked}.
Techniques such as emotion2vec in speech emotion recognition \cite{ma2023emotion2vec} and wav2vec‑based speech pretraining \cite{baevski2019effectiveness} also illustrate how contrastive and reconstruction losses can be combined to learn rich representations with minimal labels.
Recent multimodal frameworks such as SLIP \cite{mu2022slip} and self‑training strategies \cite{zoph2020rethinking} further show the potential of integrating multiple supervisory signals.
Surveys on sequential transfer learning \cite{mao2020survey} and dataset size requirements \cite{el2021large} emphasize that even small, domain‑specific datasets can yield effective pretraining when coupled with appropriate self‑supervised objectives.
\begin{table*}[p]
\centering
\caption{Summary of Self‑Supervised Pre‑training Methods for Time Series and Related Domains}
\label{tab:literature_summary}
\renewcommand{\arraystretch}{1.25}
\resizebox{\textwidth}{!}{%
\scriptsize
\begin{tabular}{p{2.5cm} c p{3cm} p{6.5cm} c}
\hline
\textbf{Author(s)} & \textbf{Year} & \textbf{Method / Focus} & \textbf{Key Contribution} & \textbf{Ref.} \\
\hline
\multicolumn{5}{l}{\cellcolor{gray!15}\textbf{\textit{Self‑Supervised Pre‑training for Time Series (Sec.~\ref{sec:sspt_ts})}}} \\
\hline
Shi et al. & 2021 & Denoising + DTW Similarity & SSL pipeline for TS classification using denoising and DTW‑based similarity on UCR2015 & \cite{shi2021self} \\
Dong et al. & 2024 & TimeSiam & Siamese network reconstructing masked subsequences via past temporal contexts & \cite{dong2024timesiam} \\
Zhang et al. & 2022 & Time–Freq. Consistency & Contrastive learning aligning time and frequency domains for cross‑domain transfer & \cite{zhang2022self} \\
Zhang et al. & 2024 & SSL Taxonomy (Survey) & Taxonomy: generative, contrastive, adversarial paradigms; key challenges identified & \cite{zhang2024self} \\
Liu et al. & 2023 & Medical TS SSL (Survey) & Review of SSL for medical TS covering 43 studies (ICU mortality, sleep scoring) & \cite{liu2023self} \\
Ma et al. & 2024 & Large Pre‑trained Models (Survey) & Survey on large pre‑trained models for TS; transformer backbones and masked modeling & \cite{ma2024survey} \\
\hline
\multicolumn{5}{l}{\cellcolor{gray!15}\textbf{\textit{Pre‑training in Clinical and Multimodal Contexts (Sec.~\ref{sec:clinical_multimodal})}}} \\
\hline
King et al. & 2023 & Multimodal ICU Pretraining & Aligns ICU time series with clinical notes via contrastive and masked token prediction & \cite{king2023multimodal} \\
Raghu et al. & 2022 & Contrastive Multimodal TS & Contrastive pre‑training integrating ECG with labs/vitals for mPAP and mortality & \cite{raghu2022contrastive} \\
Xu et al. & 2023 & TransEHR & SSL transformer for asynchronous EHR via masked value prediction and event forecasting & \cite{xu2023transehr} \\
Chowdhury et al. & 2023 & PrimeNet & Time‑sensitive contrastive learning with duration‑based masking for irregular TS & \cite{chowdhury2023primenet} \\
Malhotra et al. & 2017 & TimeNet & SSL reconstruction and contrastive losses for few‑shot transfer in clinical/sensor data & \cite{malhotra2017timenet} \\
\hline
\multicolumn{5}{l}{\cellcolor{gray!15}\textbf{\textit{Scaling Pre‑training Across Domains (Sec.~\ref{sec:scaling_domains})}}} \\
\hline
Prabhakar et al. & 2024 & LPTM & Adaptive segmentation during masked modeling for multi‑domain datasets & \cite{prabhakar2024large} \\
Dong et al. & 2023 & SimMTM & Series‑wise contrastive alignment with masked reconstruction across tasks & \cite{dong2023simmtm} \\
Zhang et al. & 2024 & UniMTS & Unifies motion TS with semantic text for zero‑shot and few‑shot performance & \cite{zhang2024unimts} \\
\hline
\multicolumn{5}{l}{\cellcolor{gray!15}\textbf{\textit{Insights from Other Modalities (Sec.~\ref{sec:other_modalities})}}} \\
\hline
Zhou; Li; Tu; Das; Rohrich et al. & Various & Masked AE for Medical Imaging & Self‑pretraining on target dataset mitigates domain shift for segmentation/classification & \cite{zhou2023self,li2023robust,tu2022self,das2025self,rohrich2025masked} \\
Ma et al. & 2023 & emotion2vec & SSL speech emotion recognition combining contrastive and reconstruction losses & \cite{ma2023emotion2vec} \\
Baevski et al. & 2019 & wav2vec Pretraining & Contrastive speech pretraining learning representations with minimal labels & \cite{baevski2019effectiveness} \\
Mu et al. & 2022 & SLIP & Multimodal framework integrating multiple supervisory signals & \cite{mu2022slip} \\
Zoph et al. & 2020 & Self‑Training Strategies & Demonstrates potential of self‑training with multiple supervisory signals & \cite{zoph2020rethinking} \\
Mao et al. & 2020 & Seq. Transfer Learning (Survey) & Survey on effective pretraining even with small domain‑specific datasets & \cite{mao2020survey} \\
El et al. & 2021 & Dataset Size Requirements & Study on dataset size requirements for effective SSL pretraining & \cite{el2021large} \\
\hline
\end{tabular}%
}
\end{table*}
\subsection{Rationale}
To the best of our knowledge, no prior work has systematically examined Self-PreTraining (SPT) in the context of medical time series. With this study, we aim to fill this gap by pursuing the following objectives:
\begin{itemize}
\item Compare transformer models trained from scratch with those initialized through SPT to quantify the benefits of this method.
\item Evaluate SPT in the simplest setting, where only univariate time-series inputs are available.
\item Extend the analysis to multivariate time series using four distinct masking strategies designed to probe different aspects of temporal and cross-channel structure.
\item Assess whether the scale of the model influences the effectiveness of SPT, particularly for deeper transformer architectures.
\item Provide an optimization-based interpretation to explain the observed performance differences between training regimes.
\end{itemize}

\section{Materials}
Table~\ref{tab:dataset_summary} summarizes the principal characteristics of the datasets considered in this work, highlighting their domain of application, classification task, cohort size, sensing modalities, segmentation strategy, evaluation protocol, and class distribution. The selected datasets cover distinct yet complementary domains, allowing for a comprehensive assessment across multimodal biomechanical, physiological, and neurological time-series data.

Comprehensive methodological details, including acquisition protocols and preprocessing procedures, are described in the preceding subsections dedicated to each dataset.
\begin{table*}[htbp]
\centering
\caption{Summary of datasets used in this study.}
\label{tab:dataset_summary}

\begin{tabular}{l p{3cm} p{2.0cm} c p{3.0cm} p{1.2cm} p{1.3cm} p{1.3cm}}
\hline
Dataset & Domain & Task & \#Subj. & Modalities & Window & Eval. & Class Dist. \\
\hline

CAMARGO 2021 \ref{subsec:camargo2021}
& Biomechanical / Multimodal Gait 
& Locomotion Classification 
& 21 
& 4 IMUs + 11 EMG (35 signals) 
& 100 ms 
& LOSO
& 20$\pm$2.5\% \\

Non-EEG \ref{subsec:stress}
& Wearable Physiological / Stress 
& Stress Detection 
& 20 
& EDA, Temp., Acc., HR, SpO$_2$ 
& 10 s 
& LOSO 
& 60\% / 40\% \\
 
Gait PD  \ref{subsec:parkinson}
& Neurological / Gait Analysis 
& PD Detection 
& 166 
& 16 Force + 2 Sum Signals 
& 5 s 
& Subject-wise 
& 69\% / 31\% \\

\hline
\end{tabular}

\end{table*}

\subsection{CAMARGO 2021}
\label{subsec:camargo2021}
This study makes use of the publicly available CAMARGO dataset \cite{camargo2021comprehensive}, a multimodal collection of sensor recordings from 21 participants. Each subject wore four IMUs located on the trunk, thigh, shank, and foot, along with 11 EMG sensors placed over key lower-limb muscles: gastrocnemius medialis, tibialis anterior, soleus, vastus medialis, vastus lateralis, rectus femoris, biceps femoris, semitendinosus, gracilis, gluteus medius, and the right external oblique.
Participants performed multiple trials across five locomotion tasks: level walking, ramp ascent and descent, and stair ascent and descent. Stair trials were completed at four step heights (\SI{102}{\milli\meter}, \SI{127}{\milli\meter}, \SI{152}{\milli\meter}, \SI{178}{\milli\meter}), while ramp trials were conducted at six inclines (\ang{5.2}, \ang{7.8}, \ang{9.2}, \ang{11}, \ang{12.4}, and \ang{18}).
Eleven EMG sensors positioned on the right side targeting major lower-limb muscle groups, and four IMUs mounted on the torso and lower limb segments to capture tri-axial acceleration and angular velocity. Altogether, the dataset contains 35 synchronized signals collected over approximately 40 minutes per subject, providing high-resolution recordings across diverse locomotion conditions. For further details on the data collection procedures and precise sensor placements, refer to the original publication \cite{camargo2021comprehensive}.
In this work, we use the dataset exactly as described above, and apply a rolling-window segmentation strategy to prepare the signals for model training. Continuous IMU and EMG recordings are divided into non-overlapping 100 ms windows, yielding roughly 5,000 samples per subject (with slight variation in both sample count and class distribution across participants). On average, each subject shows a class probability of 
20±2.5\%. The same window length and segmentation procedure are used consistently across all training configurations pre-training, fine-tuning, and training from scratch to ensure comparability of the resulting models.

\subsection{Non-EEG Dataset for Assessment of Neurological Status}
\label{subsec:stress}
This work also incorporates the Non-EEG Dataset for Assessment of Neurological Status (v1.0.0), a publicly available collection hosted on PhysioNet. The dataset includes recordings from 20 healthy adult participants, acquired by the Quality of Life Laboratory at the University of Texas at Dallas. Each subject was monitored using a set of non-EEG physiological sensors that measured electrodermal activity (EDA), skin temperature, three-axis wrist acceleration, heart rate (HR), and arterial oxygen saturation (SpO$_2$). All data are provided in WFDB format, with each participant having one file containing EDA, temperature, and accelerometry, and a second file containing HR and SpO$_2$ signals; an accompanying annotation file specifies the timing and labels of experimental phases.

Recordings were collected across seven consecutive stages designed to elicit varying physical and emotional responses. Participants began with a 5-minute resting period, followed by a physical-stress block that included standing, slow walking at 1 mph, and faster walking or jogging at 3 mph. After a second resting interval, subjects completed a cognitive-stress segment involving serial subtraction and a Stroop test, followed again by a relaxation period. An emotional-stress segment then took place, consisting of a brief anticipation interval and a 5-minute horror-movie clip, before concluding with a final resting phase. This structure yields continuous multimodal physiological data spanning relaxation, physical exertion, cognitive load, and emotional stimuli. For full methodological details, including sensor specifications and experimental protocol, refer to the original publication by \cite{birjandtalab2016non}.
For the stress dataset, windowing is performed within a leave-one-subject-out (LOSO) evaluation framework, where windows are generated only after selecting the held-out subject to avoid any form of inter-subject leakage. Each recording is segmented into temporal windows of 10 seconds, capturing sufficiently long physiological responses to characterize transitions between relaxed and stressed states. This process produces a substantial number of windowed samples per participant. The resulting class distribution remains inherently imbalanced—typically with a higher proportion of non-stress windows than stress windows reflecting the natural occurrence of these states within the experimental protocol. No artificial class balancing or resampling is applied during window creation.
The resulting class distribution remains naturally imbalanced—approximately 60\% control and 40\% Stress’s in our windowed dataset reflecting the underlying proportions of the two groups. No artificial class balancing or reweighting is applied during window generation.
\subsection{Gait in Parkinson's Disease}
\label{subsec:parkinson}
In this study we employ the publicly accessible Gait in Parkinson’s Disease v1.0.0 dataset from MIT Laboratory for Computational Physiology / PhysioNet. The dataset comprises multichannel recordings of foot‐ground reaction forces collected from 93 patients diagnosed with idiopathic Parkinson’s Disease (mean age 66.3 years, 63 \% male) and 73 healthy control subjects (mean age 66.3 years, 55 \% male). 
Under each foot of every participant, eight force sensors (Ultraflex Computer Dyno Graphy, Infotronic Inc.) were placed, yielding 16 individual sensor outputs in addition to two combined sum signals (one per foot). These signals were digitized at a sampling rate of 100 samples per second. 
The participants walked on level ground at their self‐selected pace for approximately two minutes. The available data facilitate analysis of stride‐to‐stride dynamics and variability, such as centre‐of‐pressure trajectories and timing measures (e.g., stride and swing times). 
In addition to the force sensor data, demographic information and measures of disease severity (via the Hoehn \& Yahr scale and/or the Unified Parkinson’s Disease Rating Scale) are included in the dataset. A subset of the recordings also contains dual-task walking (walking while performing serial subtraction) to explore gait under cognitive load \cite{goldberger2000physiobank}.
For the Gait in Parkinson’s Disease dataset, windowing is performed after the subject-wise train/validation/test split to prevent cross-subject information leakage. Each recording is segmented into temporal windows of 500 samples (corresponding to 5 s at the dataset’s 100 Hz sampling rate). This procedure yields a large collection of windowed gait segments for each participant. The resulting class distribution remains naturally imbalanced approximately 69\% control and 31\% Parkinson’s in our windowed dataset reflecting the underlying proportions of the two groups. No artificial class balancing or reweighting is applied during window generation.

\section{Methods}
This section outlines the architecture of the Transformer models adopted in our study used respectively for Self-Supervised Pre-Training (SPT), subsequent fine-tuning, and training from scratch—and describes the four-fold masking strategy employed for multivariate time-series learning. All three training paradigms rely on the same backbone. We choose this architecture because is the one used in the paper \cite{amos2023never} on which they showed of on a syntetic dataset the transformer was the architecture that was taking more benefit from SPT. During SPT, the model learns to reconstruct masked segments of the input sequence, enabling it to internalize modality-specific dynamics, inter-sensor correlations, and shared temporal patterns without requiring labels. Fine-tuning then adapts this pre-trained representation to the downstream classification task, while training from scratch uses identical architecture but learns all parameters solely from supervised data.

The four-fold masking strategy plays a central role in shaping the representation learned during SPT. Specifically, the time series is perturbed along multiple dimensions temporal spans, channel subsets, point-wise samples, and structured contiguous blocks allowing the model to experience a diverse set of information-removal patterns. This multi-masking design forces the encoder to infer missing information from both temporal context and cross channel interactions, which is particularly beneficial for multivariate physiological data where dependencies unfold across different timescales and sensor modalities.By maintaining an identical Transformer backbone across all training regimes, we isolate the effect of initialization and eliminate architectural confounders, ensuring a controlled and fair comparison between SPT and training from scratch. Furthermore, the unified masking interface allows different structural perturbations to be studied without altering the optimization objective or model capacity. Together, this design preserves the full expressive power of the Transformer while encouraging the learning of temporally and cross-modally coherent representations, ultimately enhancing generalization in data-limited medical settings.
\subsection{Transformer Architecture}

\begin{figure}[htbp]
  \centering
  \includegraphics[width=1.0\linewidth]{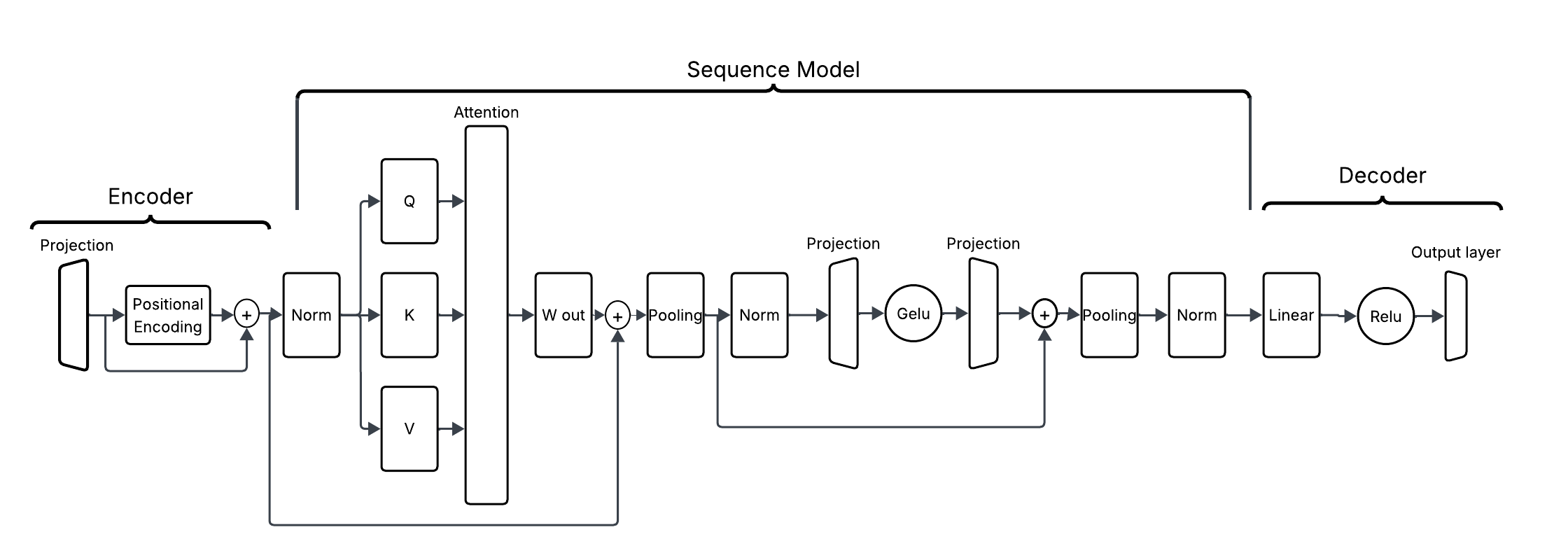}%
  \caption{Transformer architecture}
  \label{fig:TransformerArc}
\end{figure}

Figure~\ref{fig:TransformerArc} depicts the Transformer architecture employed in this study. The input consists of a univariate or multivariate time series 
\( X \in \mathbb{R}^{n \times f} \), where \( n \) denotes the sequence length (window size) and \( f \) the number of channels (with \( f=1 \) in the univariate case). The model processes each window independently.

\textbf{Projection.} The leftmost projection block represents a learnable linear mapping 
\( \mathbb{R}^{f} \rightarrow \mathbb{R}^{d} \) applied at each time step, embedding the raw sensor measurements into a \( d \)-dimensional latent space. This aligns heterogeneous sensor scales and provides a shared representational space for attention.

\textbf{Positional Encoding.} Since self-attention is permutation invariant, a positional encoding is added elementwise to the projected embeddings. This injects temporal order information, enabling the model to distinguish between different time indices within the sequence.

\textbf{Norm.} The normalization blocks denote layer normalization, stabilizing the feature distribution before attention and feed-forward transformations.

\textbf{Q, K, V (Attention Inputs).} The blocks labeled \( Q \), \( K \), and \( V \) represent learned linear projections that map the normalized embeddings into Query, Key, and Value spaces. For each time step, attention weights are computed via scaled dot-product similarity between queries and keys, determining how strongly each position attends to others.

\textbf{Attention.} The vertical attention block aggregates information across all time steps through multi-head self-attention. This operation enables the model to capture long-range temporal dependencies and cross-channel interactions within the window. Multiple heads allow the model to attend to different temporal patterns simultaneously.

\textbf{W out.} The output projection \( W_{\text{out}} \) combines the concatenated attention heads back into the model dimension \( d \). The residual connection (indicated by the \textbf{+} symbol) adds the block input to its output, improving gradient flow and preserving previously encoded information.

\textbf{Pooling.} The pooling blocks perform temporal aggregation, compressing the sequence level representations into a more compact form. This reduces dimensionality while preserving salient temporal information and prepares the representation for subsequent transformations.

\textbf{Feed Forward Projections + GELU.} The two projection blocks surrounding the GELU activation correspond to the position wise feed forward network. The first linear layer expands the embedding dimension, increasing representational capacity; GELU introduces non linearity; the second projection restores the original dimensionality. The residual connection again ensures stable optimization.

\textbf{Sequence Model Bracket.} The large bracket labeled “Sequence Model” encompasses the stacked attention and feed-forward sublayers that collectively transform the input sequence into a contextualized latent representation. Within this block, each time step representation becomes aware of the entire window.

\textbf{Decoder.} The decoder head consists of a linear transformation followed by a ReLU activation and a final output layer. During Self-PreTraining (SPT), this head predicts the masked input values (reconstruction objective). During fine-tuning or training from scratch, it maps the pooled latent representation to class logits for supervised classification.

In summary, the architecture transforms an input time series of length \( n \) into a context-aware latent embedding through repeated attention-based interactions across time. The model simultaneously captures intra-channel temporal structure and inter-channel dependencies, while residual pathways and normalization maintain stable optimization across pre-training and supervised regimes.

\subsection{Masking Policy for SPT}
\label{sec:masking_policy}
While all masking strategies remove approximately the same proportion of input elements, they differ fundamentally in their structural organization. The key distinction lies in the pattern according to which information is removed across time and feature dimensions. By varying this structure, we impose different inductive biases on the model and encourage it to learn complementary aspects of the underlying temporal data.

Unstructured masking (e.g., point-wise) removes isolated scalar values, promoting local interpolation and distributed representations. In contrast, structured masking along the temporal axis (e.g., block-wise removal) introduces contiguous gaps, forcing the model to reason over longer temporal contexts and capture global sequence dynamics. Similarly, structured masking along the feature axis (column-wise masking) removes entire sensor channels, requiring the model to infer missing modalities from cross-channel dependencies.

Thus, although the overall masking ratio remains constant, the geometric arrangement of the masked entries determines whether the model primarily learns short-range continuity, long-range temporal coherence, cross-modal redundancy, or a combination of these factors. Exploring multiple masking patterns allows us to systematically analyze how different structural corruptions influence representation learning during Self-PreTraining.

\begin{figure}[htbp]
  \centering
  \includegraphics[width=0.7\linewidth]{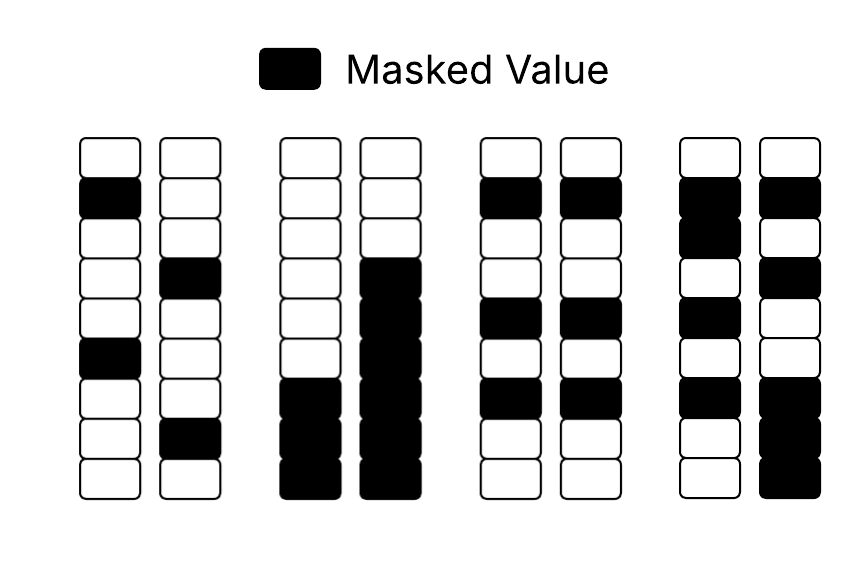}
  \caption{Masking strategies used during SPT. From left to right: point-wise masking, timestep masking, feature masking, and their mixed union.}
  \label{fig:MaskingStrategy}
\end{figure}

\paragraph{Masking Strategies.}
Figure~\ref{fig:MaskingStrategy} illustrates the masking configurations. Each strategy produces a mask with identical shape but different structural properties, thereby imposing different inductive biases.

\textbf{Point-wise masking.}  
In this configuration, each scalar entry is independently masked with probability p
This produces a fine-grained, unstructured occlusion pattern. The model must reconstruct missing values using both short range temporal context and cross feature information. This strategy resembles classical denoising autoencoding and encourages distributed, locally coherent representations.

\textbf{Column masking.}  
In this configuration, entire feature channels are removed from the input sequence. For each sample, we randomly select a fraction p of the available features and mask them across all timesteps. In other words, once a feature is selected, its entire temporal trajectory is hidden for that sequence.

This means that the model cannot rely on that sensor at any time point and must reconstruct its values using information from the remaining channels. As a result, feature masking encourages the model to learn cross-channel relationships and capture dependencies between different sensors or modalities. This is particularly relevant in multivariate physiological data, where signals often exhibit redundancy or coordinated behavior across channels.

\textbf{Block masking.}  
Block masking removes contiguous temporal segments across all features. Instead of selecting isolated timesteps, we iteratively sample starting indices and mask blocks of length $L$ along the time axis until approximately $p \cdot t$ timesteps are covered. This concentrates supervision on longer structured gaps and requires the model to infer extended temporal dynamics rather than isolated points.

\textbf{Mixed Masking Policy.}
The mixed strategy combines multiple masking axes by taking the logical union of independently sampled masks:
\[
M^{\text{mix}} 
= 
M^{\text{point}}(p_1)
\;\vee\;
M^{\text{column}}(p_2)
\;\vee\;
M^{\text{feature}}(p_3).
\]
The final mask $M^{\text{mix}}$ contains all positions masked by any individual strategy. The individual rates $p_1, p_2, p_3$ are chosen such that the expected overall masking proportion is approximately equal to the target ratio $p$. 
This mixed policy simultaneously introduces fine-grained perturbation, structured temporal gaps, and cross-channel occlusion. As a result, the model must reconstruct information across multiple axes of variation, encouraging richer and more generalizable latent representations.

\section{Experimental Setup}

\paragraph{Evaluation Protocol.}
For the CAMARGO dataset, which is formulated as a multiclass locomotion classification task, we adopted a Leave-One-Subject-Out (LOSO) cross-validation strategy. In each fold, data from $N-1$ subjects were used for training, and the remaining subject was used exclusively for testing. This procedure was repeated once per subject to obtain subject-independent performance estimates.

The Non-EEG and Parkinson’s Disease datasets correspond to binary classification problems (stress vs.\ control, and Parkinson’s vs.\ healthy control, respectively). For the Non-EEG dataset, we similarly applied LOSO.

For the Parkinson’s Disease dataset, LOSO was not feasible due to dataset organization and sample distribution constraints. Instead, we performed repeated subject-wise splits, ensuring that all samples from a given subject were assigned exclusively to either training or testing. This splitting procedure was repeated multiple times with different random partitions to obtain robust performance estimates.

\paragraph{Performance Metrics.}
Performance was evaluated using accuracy, precision, recall, and F1-score. For the multiclass CAMARGO dataset, metrics were computed using macro-averaging. For the binary datasets (Non-EEG and Parkinson’s Disease), metrics were computed in their standard binary formulation.

\paragraph{Training Configuration.}
All experiments were conducted on an NVIDIA A100 GPU with 80\,GB of memory.

For the training-from-scratch setup, models were trained for 10 epochs. A learning-rate sweep over of $0.01$ was.

For the Self-PreTraining (SPT) phase, models were pre-trained for 100 epochs using a learning rate of $0.001$, selected empirically for stable convergence. During fine-tuning, models were optimized for 15 epochs with a learning rate of $0.01$.

Early stopping was applied by monitoring the validation loss. Training was terminated if no improvement was observed for 100 consecutive epochs. In practice, convergence occurred well before reaching this limit, and the runs completed within the scheduled number of epochs.

\paragraph{Statistical Analysis.}
To assess whether performance differences between SPT and training-from-scratch were statistically significant, we employed the Wilcoxon signed-rank test with Bonferroni correction to account for multiple comparisons~\cite{gehan1965generalized,napierala2012bonferroni}.

\section{Results and Discussion}

\subsection{Univariate Analysis}
\paragraph{Experimental Design.}
The univariate experiments evaluate the impact of Structured Pre-Training (SPT) when models are trained on single-sensor modalities. For each dataset, we selected the most informative individual sensor channel based on prior domain knowledge and preliminary performance screening. The objective of this analysis is twofold: (i) to isolate how SPT interacts with individual signal characteristics, and (ii) to assess how model depth influences performance in the absence of multimodal fusion.

\begin{figure}[htbp]
  \centering
  \includegraphics[width=1.0\linewidth]{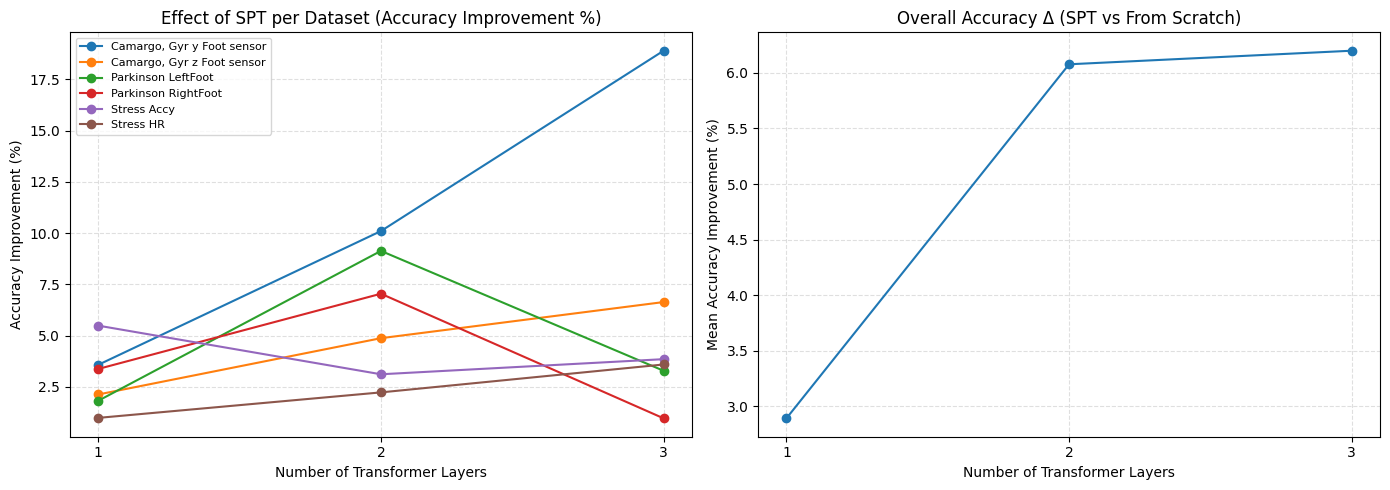}%
  \caption{Difference between From Scratch Training and SPT initialization withrespect model depth}
  \label{fig:SPT_Plot_Univariate}
\end{figure}

For each selected sensor, we trained Transformer models with 1, 2, and 3 Sequence Model layers under two initialization regimes: (i) training from scratch using Xavier initialization, and (ii) SPT initialization followed by fine-tuning. Performance metrics are reported in Table~\ref{tab:spt_benchmark}, while Figure~\ref{fig:SPT_Plot_Univariate} illustrates the depth-dependent improvement trends.

\paragraph{Sensor Selection.}
The chosen sensors correspond to those exhibiting the strongest standalone discriminative capability within each dataset:

\begin{itemize}
    \item \textbf{CAMARGO dataset:} Gyroscope signals from the foot-mounted IMU (Y-axis and Z-axis). These axes capture rotational components of gait dynamics, with the Y-axis typically encoding stronger lateral movement patterns.
    \item \textbf{Non-EEG Stress dataset:} Accelerometer (AccY) and Heart Rate (HR) signals. Accelerometer data capture agitation-related movement bursts, whereas HR reflects slower physiological stress responses.
    \item \textbf{Parkinson dataset:} LeftFoot and RightFoot force-derived gait signals, which encode stride-to-stride irregularities associated with Parkinsonian motor impairment.
\end{itemize}

These modalities were selected because they represent the strongest unimodal predictors for their respective tasks, thereby providing a meaningful testbed for assessing SPT in isolation.

\paragraph{Dataset-Specific Observations.}

\textbf{CAMARGO (Gait – Multiclass).}
The CAMARGO dataset exhibits periodic gait structure, which naturally aligns with the temporal modeling capacity of Transformers. From-scratch performance remains moderate, reflecting the multiclass complexity of locomotion tasks. SPT consistently improves performance, particularly for the Y-axis gyroscope signal, where gains increase with model depth. The structured temporal nature of gait likely allows SPT to initialize attention layers in alignment with recurring movement motifs, explaining the substantial improvements observed at two and three layers.

\textbf{Stress (Binary Classification).}
For the accelerometer-based stress dataset, baseline performance is already relatively strong due to clear motion-related stress patterns. SPT yields consistent but moderate improvements (typically 2–4\%), particularly at shallow depths. This suggests that SPT primarily stabilizes early optimization in moderately structured signals.

In contrast, the HR-based stress dataset presents smoother and more weakly structured temporal dynamics. Here, improvements are smaller (1–2.5\%) and become more apparent at higher depths. The limited temporal landmarks in HR signals constrain the representational benefit of SPT, though performance remains consistently non-degrading.

\textbf{Parkinson (Binary Classification).}
The Parkinson datasets display a distinctive behavior. Unlike other datasets, models trained from scratch benefit substantially from increased depth. Accuracy improves markedly from one to three layers even without SPT initialization. This suggests that Parkinson gait signals contain sufficiently rich temporal irregularities that deeper Transformers can exploit even under random initialization.

Nevertheless, SPT still provides significant improvements at intermediate depth (two layers), particularly for the LeftFoot dataset, where gains exceed 6\% with high statistical significance. At three layers, improvements diminish, indicating that increased depth alone already captures much of the relevant temporal structure. In this setting, SPT primarily accelerates convergence and stabilizes optimization rather than fundamentally altering the representational capacity.

\paragraph{Depth-Dependent Behavior.}
Figure~\ref{fig:SPT_Plot_Univariate} shows that the average improvement provided by SPT increases from one to two layers and stabilizes thereafter. This suggests that SPT interacts synergistically with moderate model depth: shallow models lack sufficient capacity to fully exploit structured initialization, while deeper models already capture temporal dependencies effectively, reducing marginal gains.

Importantly, SPT never degrades performance relative to training from scratch. Its benefit is most pronounced in signals with strong periodic or quasi-periodic temporal structure (e.g., gait), moderate in movement-based stress signals, and smaller in physiologically smoother signals such as heart rate. These findings indicate that SPT acts as a structured inductive prior whose impact scales with both signal regularity and model capacity.
\begin{table*}[http]
\centering
\caption{From-scratch vs. SPT performance per dataset and transformer depth, including percentage improvements. Statistical significance (Computed on the accuracy) is denoted by asterisks, with * indicating p $\leq$ 0.05, ** indicating p $\leq$ 0.01, *** indicating
p $\leq$ 0.001, and **** indicating p $\leq$ 0.0001.}
\label{tab:spt_benchmark}
\setlength{\tabcolsep}{5pt}
\renewcommand{\arraystretch}{1.2}
\begin{tabular}{l c cccc cccc cccc}
\toprule
\multirow{2}{*}{Dataset} & \multirow{2}{*}{Layers} &
\multicolumn{4}{c}{From Scratch} &
\multicolumn{4}{c}{SPT} &
\multicolumn{4}{c}{Improvement (\%)} \\
\cmidrule(lr){3-6}\cmidrule(lr){7-10}\cmidrule(lr){11-14}
& & Acc. & F1 & Recall & Spec. & Acc. & F1 & Recall & Spec. & $\Delta$Acc & $\Delta$F1 & $\Delta$Rec & $\Delta$Spec \\
\midrule

\multirow{3}{*}{Camargo, Gyr y Foot sensor}
 & 1 & 0.532$\pm$0.04 & 0.533 & 0.521 & 0.538 
     & 0.551 & 0.557 & 0.560 & 0.551
     & 1.9 * & 2.4 & 3.9 & 1.3 \\
 & 2 & 0.525$\pm$0.05 & 0.532 & 0.522 & 0.527
     & 0.578 & 0.582 & 0.579 & 0.572
     & 5.3*** & 4.7 & 5.4 & 4.5 \\
 & 3 & 0.524$\pm$0.03 & 0.527 & 0.524 & 0.523
     & 0.623 & 0.629 & 0.617 & 0.620
     & 9.9**** & 10.2 & 9.3 & 9.7 \\
\midrule

\multirow{3}{*}{Camargo, Gyr z Foot sensor}
 & 1 & 0.565$\pm$0.03 & 0.547 & 0.566 & 0.545
     & 0.577 & 0.575 & 0.578 & 0.578
     & 1.2 & 2.8 & 1.2 & 3.2 \\
 & 2 & 0.575$\pm$0.03 & 0.574 & 0.578 & 0.570
     & 0.603 & 0.598 & 0.596 & 0.607
     & 2.8* & 2.4 & 1.8 & 3.7 \\
 & 3 & 0.573$\pm$0.04 & 0.569 & 0.566 & 0.569
     & 0.611 & 0.610 & 0.614 & 0.605
     & 3.8** & 4.1 & 4.8 & 3.6 \\
\midrule

\multirow{3}{*}{Stress Accy}
 & 1 & 0.748$\pm$0.03 & 0.747 & 0.748 & 0.725
     & 0.789 & 0.796 & 0.783 & 0.772
     & 4.1** & 4.9 & 3.5 & 4.6 \\
 & 2 & 0.772$\pm$0.02 & 0.781 & 0.775 & 0.754
     & 0.796 & 0.800 & 0.795 & 0.784
     & 2.4 & 1.9 & 2.3 & 3.0 \\
 & 3 & 0.780$\pm$0.03 & 0.793 & 0.781 & 0.778
     & 0.810 & 0.816 & 0.808 & 0.800
     & 3.0* & 2.3 & 2.8 & 2.2 \\
\midrule
                    
\multirow{3}{*}{Stress HR}
 & 1 & 0.713$\pm$0.03 & 0.730 & 0.718 & 0.708
     & 0.720 & 0.738 & 0.7231 & 0.713
     & 0.7 & 0.8 & 0.5 & 0.5 \\
 & 2 & 0.719$\pm$0.06 & 0.724 & 0.716 & 0.711
     & 0.735 & 0.744 & 0.73 & 0.723
     & 1.6 & 2.0 & 1.4 & 1.2 \\
 & 3 & 0.697$\pm$0.05 & 0.698 & 0.683 & 0.690
     & 0.722 & 0.7280 & 0.718 & 0.717
     & 2.5* & 3.0 & 3.6 & 2.8 \\
\midrule
\bottomrule

\multirow{3}{*}{Parkinson LeftFoot}
 & 1 & 0.715$\pm$0.04 & 0.657 & 0.705 & 0.713
     & 0.728 & 0.700 & 0.712 & 0.729
     & 1.3 & 4.3 & 0.7 & 1.6 \\
 & 2 & 0.745$\pm$0.03 & 0.738 & 0.740 & 0.745
     & 0.813 & 0.809 & 0.813 & 0.818
     & 6.8**** & 7.1 & 7.3 & 7.3 \\
 & 3 & 0.823$\pm$0.01 & 0.815 & 0.819 & 0.822
     & 0.850 & 0.848 & 0.857 & 0.854
     & 2.7 & 3.3 & 3.8 & 3.2 \\
\midrule

\multirow{3}{*}{Parkinson RightFoot}
 & 1 & 0.742$\pm$0.01 & 0.724 & 0.730 & 0.741
     & 0.767 & 0.739 & 0.763 & 0.764
     & 2.5 & 1.5 & 3.3 & 2.3 \\
 & 2 & 0.753$\pm$0.03 & 0.731 & 0.751 & 0.757
     & 0.806 & 0.789 & 0.805 & 0.801
     & 5.3*** & 5.8 & 5.4 & 4.4 \\
 & 3 & 0.834$\pm$0.02 & 0.831 & 0.837 & 0.831
     & 0.842 & 0.839 & 0.841 & 0.842
     & 0.8 & 0.8 & 0.4 & 1.1 \\
\midrule
\bottomrule

\end{tabular}
\end{table*}

\newpage

\subsection{Multivariate Analysis}

\paragraph{Experimental Design.}
The multivariate experiments evaluate the Transformer when trained jointly on all available sensor channels, allowing the model to capture both temporal dynamics and cross-channel dependencies. For each dataset (Camargo (Entire Foot Sensor), Stress (Accy + HR), and Parkinson (Right Force + Left Force)), we compare training from scratch (Xavier initialization) with four SPT variants: \textit{SPT-Point}, \textit{SPT-Block}, \textit{SPT-Column}, and \textit{SPT-Mixed}. Models are evaluated at depths of 1, 2, and 3 encoder layers.

During pre-training, all SPT models are optimized using a \textbf{Masked Mean Squared Error (Masked MSE)} loss, computed exclusively over masked elements. This ensures that the model is explicitly trained to reconstruct missing values rather than copying visible inputs. Downstream fine-tuning uses supervised cross-entropy loss.

The quantitative results are reported in Table~\ref{tab:spt_benchmark_multi}.

\paragraph{Global Observations.}
Across all three datasets, SPT variants consistently outperform training from scratch. However, the magnitude and scaling of improvements depend on:
(i) the intrinsic temporal structure of the task,
(ii) the redundancy and synchronization across channels,
(iii) the structural bias imposed by the masking strategy.

Camargo and Parkinson exhibit strong periodic structure (gait cycles), while Stress contains heterogeneous physiological signals (EDA, HR, accelerometry) with weaker cross-channel alignment. These structural differences explain the differentiated impact of each masking policy.

\subsubsection*{SPT-Point}

SPT-Point applies independent masking at the scalar level. It resembles multivariate denoising autoencoding and primarily reinforces robustness to local perturbations.

In Table~\ref{tab:spt_benchmark_multi}, SPT-Point yields consistent but moderate improvements. The gains are particularly visible in Camargo (depth 2: 0.828 → 0.848, p $\leq$ 0.05), where local periodic structure dominates but cross-channel alignment is imperfect. 

\begin{figure}[htbp]
\centering
\includegraphics[width=1.0\linewidth]{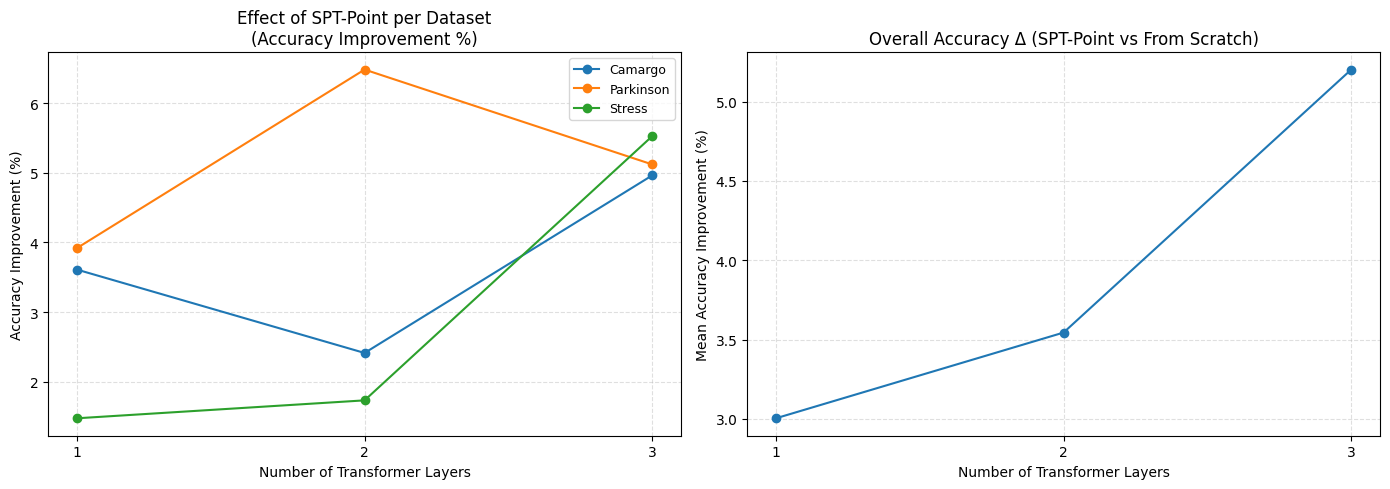}%
\caption{SPT-Point improvement vs transformer depth. Improvements grow modestly with scale.}
\label{fig:SPT_Plot_Point}
\end{figure}

However, improvements do not scale aggressively with depth, as shown in Fig.~\ref{fig:SPT_Plot_Point}. This reflects the fact that pointwise masking does not explicitly enforce structured temporal or cross-channel reasoning.

\subsubsection*{SPT-Block}

SPT-Block masks contiguous temporal segments, requiring the model to reconstruct extended time intervals. This directly encourages long-range temporal modeling.

As shown in Table~\ref{tab:spt_benchmark_multi}, SPT-Block performs particularly well in temporally structured datasets such as Parkinson. At depth 2, accuracy reaches 0.964 (p $\leq$  0.001), substantially exceeding from-scratch training.

\begin{figure}[htbp]
\centering
\includegraphics[width=1.0\linewidth]{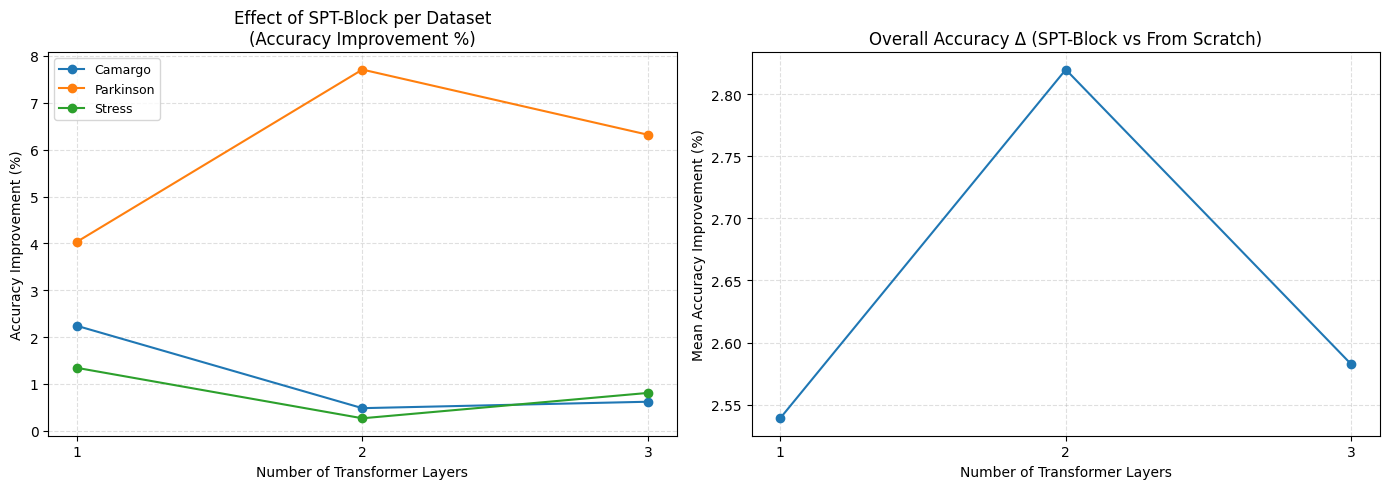}%
\caption{SPT-Block strongest improvements at mid scale, limited benefit beyond 2 layers.}
\label{fig:SPT_Plot_Block}
\end{figure}

The scaling behavior in Fig.~\ref{fig:SPT_Plot_Block} reveals that improvements peak at intermediate depth. This suggests that block masking efficiently encodes dominant temporal structure early, but additional layers do not proportionally increase the benefit.

\subsubsection*{SPT-Column}

SPT-Column removes entire feature channels during pre-training. This forces reconstruction via cross-channel inference and explicitly promotes sensor fusion.

In Table~\ref{tab:spt_benchmark_multi}, SPT-Column demonstrates strong gains in Camargo and Parkinson, both of which contain synchronized multi-sensor measurements. At depth 3 in Camargo, accuracy reaches 0.869 (p $\leq$ 0.001).

\begin{figure}[htbp]
\centering
\includegraphics[width=1.0\linewidth]{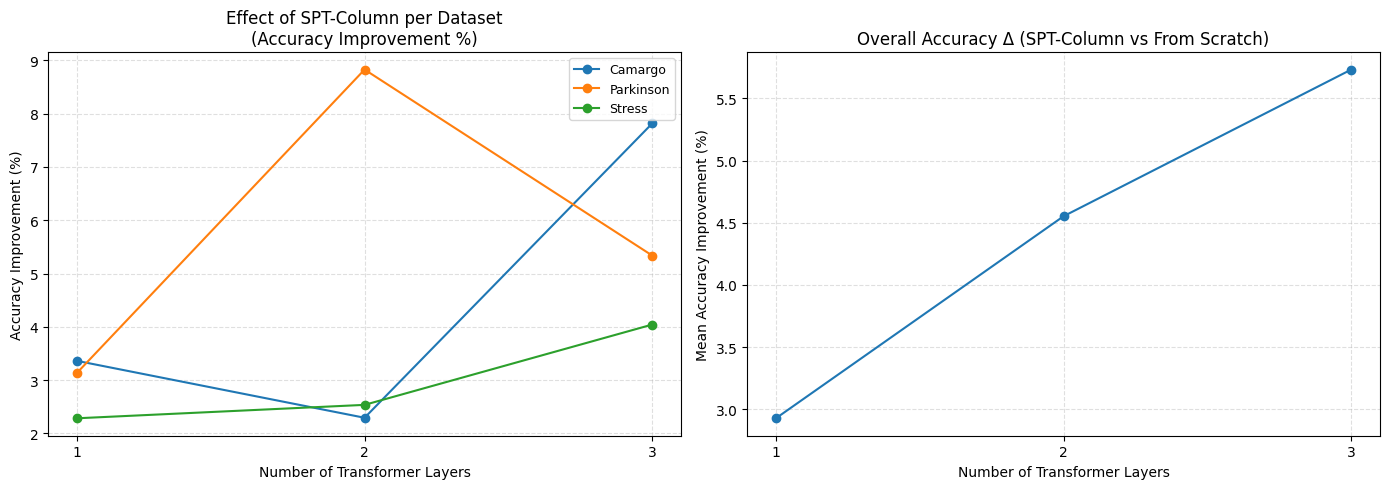}%
\caption{SPT-Column the strongest and most consistent scaling with transformer depth.}
\label{fig:SPT_Plot_Column}
\end{figure}

Figure~\ref{fig:SPT_Plot_Column} shows the strongest depth-dependent growth among all strategies. This indicates that deeper Transformers are particularly effective at exploiting structured cross-channel priors introduced by column masking.

\subsubsection*{SPT-Mixed}

SPT-Mixed combines point, block, and column masking into a unified objective. This hybrid strategy introduces uncertainty across multiple structural axes simultaneously.

As shown in Table~\ref{tab:spt_benchmark_multi}, SPT-Mixed provides the most stable improvements across heterogeneous datasets, particularly in Stress. At depth 3, accuracy reaches 0.788 (p $\leq$ 0.01), outperforming individual masking strategies.

\begin{figure}[htbp]
\centering
\includegraphics[width=1.0\linewidth]{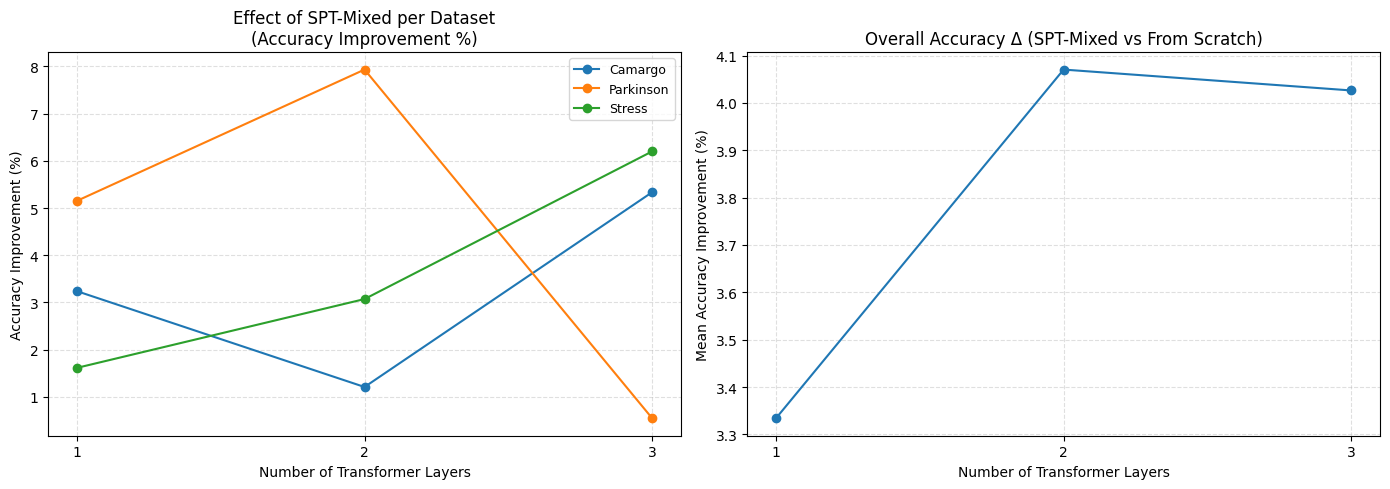}%
\caption{SPT-Mixed strongest in heterogeneous settings, scaling depends on dataset structure.}
\label{fig:SPT_Plot_Mixed}
\end{figure}

The scaling behavior in Fig.~\ref{fig:SPT_Plot_Mixed} varies by dataset. Parkinson peaks at intermediate depth (due to already strong baseline structure), while Stress continues improving with depth, reflecting its heterogeneous multimodal composition.

\paragraph{Interpretation.}

Together, these results demonstrate that the effectiveness of SPT depends on the alignment between masking structure and task structure.

\begin{itemize}
    \item In strongly periodic datasets (Parkinson, Camargo), block and column masking are particularly effective because they align with temporal and cross-channel regularities.
    \item In heterogeneous physiological datasets (Stress), mixed masking performs best because no single structural prior dominates.
\end{itemize}

As model depth increases, the performance gap between SPT and from-scratch models widens. This indicates that structured pre-training improves optimization stability and allows deeper Transformers to better exploit their representational capacity without overfitting.
\subsection{Optimization Perspective}
From an optimization perspective, SPT operates as a structured warm-start mechanism. When transformers are initialized randomly, especially with small or noisy datasets, they must discover temporal structure from scratch, often becoming trapped in poor local minima. SPT mitigates this by introducing a reconstruction-based pre-training phase that aligns the parameters toward a useful manifold. This lowers gradient variance during fine-tuning and directs attention maps toward meaningful temporal and cross-modal patterns. Importantly, SPT reduces loss landscape sharpness, a property associated with better generalization. By pre-training on a reconstruction task without labels, models converge more smoothly during downstream optimization, particularly when limited labeled examples are available — a common condition in clinical applications.

Altogether, these results show that SPT provides a consistently positive impact that scales with model depth and remains robust across downstream tasks. It emerges as a competitive alternative to external self-supervised pre-training methods, with the key advantage that it requires no additional data beyond the target dataset itself. This has critical implications for clinical AI, where labeled data are scarce and generalization across domains remains challenging.

Despite its strengths, the approach is not without limits. The degree of improvement varies based on the inherent temporal structure of the data, and the masking strategies perform differently depending on the presence of cross-channel correlations. Furthermore, the study focuses on classification tasks using fixed-length windows. Future work should investigate how SPT performs on forecasting, detection, and segmentation tasks, especially with variable-length inputs or irregularly sampled data — common characteristics in medical contexts. Exploring the full architecture design space, including width and number of attention heads, could further deepen our understanding of how SPT shapes learning dynamics.

In conclusion, SPT stands as a powerful, dataset-aligned initialization method that brings significant and consistent improvements across real-world clinical signals. It decouples temporal representation learning from label dependency, yields scalable benefits as transformer depth increases, and adapts across modalities and tasks. By structuring model initialization in ways that reflect signal modality and temporal dynamics, SPT addresses key challenges in medical machine learning, offering a highly practical pathway toward more intelligent, robust, and data-efficient clinical models.

\begin{table}[t]
\centering
\caption{Multivariate performance: from-scratch training versus SPT variants. Only accuracy
reports mean $\pm$ standard deviation across folds. Significance as in
Table~\ref{tab:spt_benchmark}. SPT-Mixed is reported separately in
Table~\ref{tab:multivariate_mixed}. Take-away: every SPT variant matches or improves on
from-scratch training in every cell, significantly so in the majority, and the best variant is
the one whose masking geometry matches the signal block and column for periodic gait, point
and mixed for heterogeneous physiology.}
\label{tab:multivariate}
\setlength{\tabcolsep}{3pt}
\renewcommand{\arraystretch}{1.0}
\resizebox{\textwidth}{!}{%
\begin{tabular}{l c cccc cccc cccc cccc}
\toprule
\multirow{2}{*}{Dataset} & \multirow{2}{*}{L} &
\multicolumn{4}{c}{From Scratch} &
\multicolumn{4}{c}{SPT-Point} &
\multicolumn{4}{c}{SPT-Block} &
\multicolumn{4}{c}{SPT-Column} \\
\cmidrule(lr){3-6}\cmidrule(lr){7-10}\cmidrule(lr){11-14}\cmidrule(lr){15-18}
 & & Acc. & F1 & Rec. & Spec. & Acc. & F1 & Rec. & Spec. & Acc. & F1 & Rec. & Spec. & Acc. & F1 & Rec. & Spec. \\
\midrule
\multirow{3}{*}{Camargo}
 & 1 & 0.803$\pm$0.066 & 0.842 & 0.816 & 0.819 & 0.832$\pm$0.056 & 0.850 & 0.832 & 0.835 & 0.821$\pm$0.059 & 0.845 & 0.834 & 0.826 & 0.830$\pm$0.050 & 0.844 & 0.826 & 0.838 \\
 & 2 & 0.828$\pm$0.072 & 0.857 & 0.821 & 0.836 & 0.848$\pm$0.059$^{*}$ & 0.865 & 0.855 & 0.851 & 0.832$\pm$0.063 & 0.857 & 0.836 & 0.841 & 0.847$\pm$0.065$^{*}$ & 0.866 & 0.848 & 0.850 \\
 & 3 & 0.806$\pm$0.075 & 0.810 & 0.806 & 0.802 & 0.846$\pm$0.054$^{**}$ & 0.862 & 0.843 & 0.851 & 0.811$\pm$0.075 & 0.816 & 0.801 & 0.808 & 0.869$\pm$0.056$^{***}$ & 0.865 & 0.849 & 0.851 \\
\midrule
\multirow{3}{*}{Stress}
 & 1 & 0.744$\pm$0.063 & 0.732 & 0.746 & 0.766 & 0.755$\pm$0.063 & 0.749 & 0.757 & 0.772 & 0.754$\pm$0.063 & 0.745 & 0.754 & 0.765 & 0.761$\pm$0.063 & 0.732 & 0.766 & 0.783 \\
 & 2 & 0.749$\pm$0.058 & 0.747 & 0.753 & 0.762 & 0.762$\pm$0.062 & 0.757 & 0.762 & 0.780 & 0.751$\pm$0.056 & 0.744 & 0.751 & 0.761 & 0.768$\pm$0.054$^{*}$ & 0.763 & 0.770 & 0.774 \\
 & 3 & 0.742$\pm$0.056 & 0.744 & 0.749 & 0.765 & 0.783$\pm$0.051$^{*}$ & 0.760 & 0.784 & 0.799 & 0.748$\pm$0.062 & 0.737 & 0.748 & 0.759 & 0.772$\pm$0.045 & 0.774 & 0.765 & 0.783 \\
\midrule
\multirow{3}{*}{Parkinson}
 & 1 & 0.893$\pm$0.047 & 0.889 & 0.883 & 0.898 & 0.928$\pm$0.025$^{*}$ & 0.926 & 0.930 & 0.918 & 0.929$\pm$0.032$^{*}$ & 0.927 & 0.930 & 0.922 & 0.921$\pm$0.044$^{*}$ & 0.922 & 0.923 & 0.920 \\
 & 2 & 0.895$\pm$0.055 & 0.893 & 0.896 & 0.888 & 0.953$\pm$0.027$^{**}$ & 0.954 & 0.947 & 0.954 & 0.964$\pm$0.041$^{***}$ & 0.968 & 0.961 & 0.957 & 0.974$\pm$0.043$^{****}$ & 0.968 & 0.971 & 0.961 \\
 & 3 & 0.918$\pm$0.032 & 0.909 & 0.914 & 0.916 & 0.965$\pm$0.041$^{**}$ & 0.959 & 0.967 & 0.964 & 0.976$\pm$0.038$^{***}$ & 0.969 & 0.977 & 0.973 & 0.967$\pm$0.026$^{**}$ & 0.962 & 0.968 & 0.959 \\
\bottomrule
\end{tabular}%
}
\end{table}

\begin{table}[t]
\centering
\caption{SPT-Mixed, multivariate. From-scratch accuracy is repeated from
Table~\ref{tab:multivariate_mixed} for reference; $\Delta$ is the gain in absolute percentage points.
Significance as in Table~\ref{tab:spt_benchmark}. Take-away: mixed masking is the strongest variant
on heterogeneous physiology, where no single structural prior dominates, but it is the least
consistent on periodic gait.}
\label{tab:multivariate_mixed}
\small
\setlength{\tabcolsep}{6pt}
\renewcommand{\arraystretch}{1.1}
\begin{tabular}{l c c ccccc}
\toprule
\multirow{2}{*}{Dataset} & \multirow{2}{*}{L} & From Scratch & \multicolumn{5}{c}{SPT-Mixed} \\
\cmidrule(lr){3-3}\cmidrule(lr){4-8}
 & & Acc. & Acc. & F1 & Rec. & Spec. & $\Delta$Acc \\
\midrule
\multirow{3}{*}{Camargo}
 & 1 & 0.803$\pm$0.066 & 0.829$\pm$0.061 & 0.852 & 0.830 & 0.833 & +2.6 \\
 & 2 & 0.828$\pm$0.072 & 0.838$\pm$0.057 & 0.858 & 0.839 & 0.842 & +1.0 \\
 & 3 & 0.806$\pm$0.075 & 0.849$\pm$0.052$^{**}$ & 0.865 & 0.847 & 0.851 & +4.3 \\
\midrule
\multirow{3}{*}{Stress}
 & 1 & 0.744$\pm$0.063 & 0.756$\pm$0.064 & 0.716 & 0.756 & 0.767 & +1.2 \\
 & 2 & 0.749$\pm$0.058 & 0.772$\pm$0.054$^{**}$ & 0.744 & 0.772 & 0.778 & +2.3 \\
 & 3 & 0.742$\pm$0.056 & 0.788$\pm$0.039$^{**}$ & 0.769 & 0.788 & 0.797 & +4.6 \\
\midrule
\multirow{3}{*}{Parkinson}
 & 1 & 0.893$\pm$0.047 & 0.939$\pm$0.05\phantom{0}$^{**}$ & 0.941 & 0.940 & 0.941 & +4.6 \\
 & 2 & 0.895$\pm$0.055 & 0.966$\pm$0.05\phantom{0}$^{****}$ & 0.967 & 0.971 & 0.963 & +7.1 \\
 & 3 & 0.918$\pm$0.032 & 0.923$\pm$0.03\phantom{0} & 0.919 & 0.926 & 0.920 & +0.5 \\
\bottomrule
\end{tabular}
\end{table}

\section{Conclusions}
In this work, we introduced Self-PreTraining (SPT) as a novel initialization strategy for transformer models applied to medical time series classification. Unlike supervised pre-training or approaches that depend on external datasets, SPT leverages only the data available in the target task through an unsupervised reconstruction objective. By masking and reconstructing segments of the input signal, SPT imposes a temporal and structural prior that helps shape the model's learning dynamics before fine-tuning.

Across diverse datasets spanning inertial gait signals, physiological stress markers, and multimodal Parkinsonian movement data SPT consistently outperforms training from scratch. These improvements are especially pronounced in deeper transformer architectures, where SPT mitigates common optimization challenges such as unstable gradient propagation and poor generalization. Moreover, our comparative analysis of masking strategies reveals that no single approach dominates universally: SPT-Point introduces localized robustness, SPT-Block captures temporal continuity, and SPT-Column exploits cross-sensor redundancy, while SPT-Mixed effectively synthesizes these benefits for heterogeneous datasets.

Overall, SPT represents a practical, domain-aligned method for enhancing transformer-based models in clinical applications. It requires no external data or architectural modifications, making it particularly suitable for scenarios with limited labeled samples a frequent constraint in medical machine learning. Future directions include broadening this framework to time series forecasting, imputation, and segmentation tasks, as well as exploring its impact on irregular or asynchronous signals. By offering a flexible and effective initialization approach, SPT has the potential to facilitate the deployment of more reliable and data efficient models in real world healthcare settings.
\section*{Acknowledgment}
Coser Omar is a Ph.D. student enrolled in the National Ph.D. in Artificial Intelligence, XXXVIII cycle, course on Health and Life Sciences, organized
by Università Campus Bio-Medico di Roma.

\bibliographystyle{IEEEtran}
\bibliography{ref}

\end{document}